\documentclass{article}
\usepackage{ijcai18}

\usepackage{graphicx} 
\usepackage{subfigure} 
\usepackage{floatrow}
\usepackage{dblfloatfix}
\usepackage{amsmath}
\usepackage{amssymb}
\usepackage{algorithm}
\usepackage{graphicx}
\usepackage{algorithmic}
\usepackage{times}
\usepackage{xcolor}
\usepackage[font={small}]{caption}
\usepackage{latexsym} 

\DeclareMathOperator*{\minA}{min} 

\title{Small-Variance Asymptotics for Nonparametric Bayesian \\ Overlapping Stochastic Blockmodels}

 \author{
 Gundeep Arora, 
 Anupreet Porwal, 
 Kanupriya Agarwal,
 Avani Samdariya,
 Piyush Rai
 \\ 
 Indian Institute of Technology, Kanpur \\
 \{gundeep, anupreet, kagarwal, savani, rpiyush\}@iitk.ac.in
 }

\begin{document}

\maketitle

\begin{abstract}
The latent feature relational model (LFRM) is a generative model for graph-structured data to learn a binary vector representation for each node in the graph. The binary vector denotes the node's membership in one or more communities. At its core, the LFRM~\cite{miller2009nonparametric} is an \emph{overlapping} stochastic blockmodel, which defines the link probability between any pair of nodes as a bilinear function of their community membership vectors.  Moreover, using a nonparametric Bayesian prior (Indian Buffet Process) enables learning the number of communities automatically from the data. However, despite its appealing properties, inference in LFRM remains a challenge and is typically done via MCMC methods. This can be slow and may take a long time to converge. In this work, we develop a small-variance asymptotics based framework for the non-parametric Bayesian LFRM. This leads to an objective function that retains the nonparametric Bayesian flavor of LFRM, while enabling us to design \emph{deterministic} inference algorithms for this model, that are easy to implement (using generic or specialized optimization routines) and are fast in practice. Our results on several benchmark datasets demonstrate that our algorithm is competitive to methods such as MCMC, while being much faster. 
\end{abstract}

\section{Introduction}

Relational data, such as graphs given as adjacency matrices, are prevalent in many domains, such as analysis of social networks, biological networks, citation networks, etc. Stochastic  blockmodels and its extensions~\cite{nowicki2001estimation,kemp2006learning,airoldi2008mixed,miller2009nonparametric,zhou2015infinite} are attractive models for such graph-structured data. These models are commonly used for discovering the underlying latent structure in the graph (e.g., via low-dimensional vector space representation of the nodes) and for link-prediction. The latent feature relational model (LFRM)~\cite{miller2009nonparametric} is a particularly attractive variant of stochastic blockmodels that allows each node to simultaneously belong to multiple communities by modeling each node via a binary membership vector. This LFRM can also be seen as learning an \emph{overlapping} clustering of nodes in the graph (each community represents a cluster). However, unlike various other models for learning overlapping clustering of nodes in a graph~\cite{xie2013overlapping}, the LFRM generative model also defines the probability of a link between any pair of node via a bilinear function of their community membership vectors. As a consequence, it can also be used for link-prediction, unlike other overlapping clustering models for graphs~\cite{xie2013overlapping}, that can only learn community memberships but are not suited for link-prediction. Another very appealing property of the LFRM is that the number of communities can be inferred from the data using an Indian Buffet Process prior~\cite{griffiths2011indian} on the binary node-community assignment matrix.

Despite the expressiveness and modeling flexibility, inference in the LFRM however remains a challenge. The model is non-conjugate and the only existing inference method is based on Markov Chain Monte Carlo (MCMC) sampling~\cite{miller2009nonparametric}. MCMC based methods can be slow to mix and converge, especially for nonparametric Bayesian models like LFRM. It is therefore highly desirable to develop faster, alternative inference methods the LFRM.

In this work, we appeal to the idea of small-variance asymptotics~\cite{kulis2011revisiting,broderick2013mad} in the context of the LFRM to get an equivalent non-probabilistic model. The resulting model retains the flavor of the original LFRM (e.g., the ability to infer the number of communities), but has a much simpler inference procedure which boils down to solving an optimization problem, for which existing off-the-shelf or specialized optimization routines can be used. We would like to note here that the idea is small-variance asymptotics (SVA) has also been explored recently to obtain non-probabilistic counterparts of various other nonparametric Bayesian models. However, unlike these recent works, which apply SVA for models of i.i.d./sequential vector-valued data~\cite{broderick2013mad,roychowdhury2013small,wang2015dp}, our work is motivated by the need of developing SVA based algorithms for relational data, such as graphs. We believe our work will motivate and open door to the design of fast, deterministic algorithms for learning from relational data. Our experiments on several benchmark datasets show that our algorithm attains improved/similar link-prediction accuracies as compared to MCMC based inference for LFRM, which being much faster.


\section{Latent Feature Relational Model}

We first introduce notation and problem setup and then briefly describe the nonparametric Bayesian latent feature relational model~\cite{miller2009nonparametric} (LFRM) for network data for which we develop the small-variance asymptotics to design the inference algorithm for LFRM.

We assume that the data is given as a graph between $N$ entities, represented as an $N\times N$ adjacency matrix $\mathbf{Y}$ where $y_{ij}=1$ denotes the presence of a link (edge) between node $i$ and node $j$, and $y_{ij}=0$ denotes that there is no link. The matrix $\mathbf{Y}$, however is only partially observed and the goal is to predict the presence/absence of edges where it is not observed. This is essentially a link-prediction task.

The LFRM~\cite{miller2009nonparametric} assumes that node $i$ in the graph is associated with a binary latent feature vector $\mathbf{z}_i \in \{0,1\}^{K^{+}}$ where $K^{+}$ denotes the total number of latent features. Note that  $K^{+}$ can also be thought of as denoting the total number of communities/clusters. Here, $z_{ik}=1$ indicates that node $i$ contains latent feature $k$, which is equivalent to saying that node $i$ belongs to community $k$ (and $z_{ik}=0$ otherwise). Note that in the LFRM, a node can potentially belong to more than one community. We represent the latent feature representation of all the entities by $\mathbf{Z}$, as the $N\times K^{+}$ binary matrix, which can also be interpreted as the node-community assignment matrix. In the rest of the exposition, we will sometimes use the terms latent feature, community, and cluster, interchangeably -- they all refer to the same. 

LFRM models the probability $p_{ij} \in (0,1)$ of a link between node $i$ and node $j$ as a bilinear function of their latent feature vectors (denoting their cluster assignments)
\begin{equation}
\label{eq:prob_lfrm}
    p_{ij} = \sigma(\mathbf{z}_i^T\mathbf{W}\mathbf{z}_j)
\end{equation}
where $\sigma(x) = \frac{1}{1+\exp(-x)}$ is the sigmoid function. Here $\mathbf{W}$ denote a real valued $K^{+}\times K^{+}$ matrix, with $w_{kk'}$ denoting the weight affecting the probability of link between node $i$ belonging to cluster $k$ and node $j$ belonging to cluster $k^\prime$.

The overall likelihood for the model can be written as 
\begin{equation}
 P(\mathbf{Y}|\mathbf{Z},\mathbf{W}) = \prod_{i, j=1}^{N} P(y_{ij}|\mathbf{z}_i,\mathbf{z}_j,\mathbf{W})
\end{equation}
where each $P(y_{ij}|\mathbf{z}_i,\mathbf{z}_j,\mathbf{W})$ is a Bernoulli with probability $p_{ij}$ as defined in Eq.~\ref{eq:prob_lfrm}. Assuming the observations to be i.i.d. conditioned on the latent features, the likelihood will be

\begin{equation}
\begin{aligned}
    P(\mathbf{Y}|\mathbf{Z},\mathbf{W}) = \prod_{i, j=1}^{N} p_{ij}^{y_{ij}}(1-p_{ij})^{1-y_{ij}}
\end{aligned}
\end{equation}



The LFRM model contains two main unknowns: the binary matrix $\mathbf{Z}$ of size $N\times K^{+}$ and the real-valued matrix $\mathbf{W}$ of size $K^{+}\times K^{+}$. The LFRM~\cite{miller2009nonparametric} assumes Gaussian priors on each entry $w_{kk'}$ in $\mathbf{W}$
\begin{equation}
    w_{kk'}\sim \mathcal{N}(0,\sigma^2_w) 
\end{equation}

In order to automatically learn the appropriate number of latent features (i.e., number of communities/clusters), LFRM posits an Indian Buffet Process (IBP) prior \cite{griffiths2011indian} on the binary matrix $\mathbf{Z}$. This non-parametric prior can be explained through a culinary metaphor, where each customer samples dishes from an infinitely long buffet dish-list. For each customer $n=1,\ldots,N$, an already sampled dish $k$ is chosen with a probability based on how many previous customers have sampled that dish. Thereafter, customer $n$ samples $\text{Poisson}(\alpha/n)$ new dishes, where $\alpha > 0$ is the IBP hyperparameter.  The subset of sampled dishes by a customer represents the binary latent feature. When considering all the customers, the process is equivalent to sampling a binary matrix whose number of columns is equal to the total number of unique dishes sampled. $N$ entities sample a total of $K^{+}$ features and $\mathbf{Z}_{1:N, 1:K^{+}}$ is the resulting feature allocation matrix. 

As shown in~\cite{griffiths2011indian}, the IBP prior on $\mathbf{Z}$ can be written as follows
\begin{equation}
    P(\mathbf{Z}) = \frac{\alpha^{K^{+}}}{\prod_{h=1}^{H} \tilde{K_{h}}!} \exp({-\sum_{n=1}^{N}} \frac{\alpha}{n}) \prod_{k=1}^{K^{+}} S_{N,k}^{-1} {\binom{N}{S_{N,k}}}^{-1}
    \label{eq:ibp}
\end{equation}

where $h$ represents the total number of unique decimal values of the $N\times 1$ binary vector $\mathbf{Z}_{1:N, k}$ across the $K$ columns of $\mathbf{Z}$ and $\tilde{K_{h}}!$ is the number of $k$ with $h^{th}$ unique value of this vector. $S_{N,k}$ denotes the count of feature k being one for first N entities which means that $n^{th}$ entity samples feature $k$ with probability ${S_{n-1,k}}/{n}$. 

With the priors on $\mathbf{W}$ and $\mathbf{Z}$ specified, we summarize the LFRM generative model~\cite{miller2009nonparametric}
 \begin{align}
     \mathbf{Z} &\sim \text{IBP}(\alpha) \\
    w_{kk'} & \sim \mathcal{N}(0,\sigma_w^2) \quad  \textrm{$\forall$ $k,k'$} \\
    y_{ij} & \sim \textrm{Bernoulli} (\sigma( \mathbf{z}_{i}^T \mathbf{W} \mathbf{z}_{j}))
 \end{align}

Exact inference in this model is intractable and MCMC based inference was proposed in~\cite{miller2009nonparametric}. Since MCMC can be slow to mix and converge, here we present a new inference algorithm, motivated by the idea of small-variance asymptotics~\cite{kulis2011revisiting,broderick2013mad} for the LFRM, which we describe next.

\section{Small-Variance Asymptotics for LFRM}

To develop the small-variance asymptotics (SVA) for the LFRM, we will take the MAP objective (the log of posterior $p(\mathbf{Z},\mathbf{W}|\mathbf{Y}$) for the model and take the small-variance limit of the objective to obtain an objective function which can be optimized w.r.t. $\mathbf{Z}$ and $\mathbf{W}$ to find point estimates of these unknowns. This construction is motivated by~\cite{broderick2013mad} who applied SVA for doing inference in linear Gaussian models with an a priori unknown number of latent features. However, while linear Gaussian models are designed for vector-valued data, our focus here is on models for relational data, such as LFRM. Moreover, while their model had a Gaussian likelihood with a natural variance term, for LFRM the likelihood is Bernoulli. To apply SVA for a model with Bernoulli likelihood, we leverage the equivalence of exponential family and Bregman divergence~\cite{jiang2012small} and represent the Bernoulli as a \emph{scaled} Bernoulli, which  will enable us to apply the SVA idea for LFRM.    

\section{Bregman Divergence and Scaled Bernoulli}

In this section, we establish the functional form of the scaled-likelihood (LFRM likelihood is Bernoulli), that can then be used to obtain the small variance asymptotics objective from the posterior, for the LFRM. To this end, we first express the Bernoulli distribution in its canonical form, using a generalised distance by incorporating the bijective relationship between Bregman divergences and exponential families, discussed in \cite{banerjee2005clustering}.
A likelihood $p(x|q) = \text{Bernoulli}(q)$, has the exponential family representation
\begin{equation}
        P(x|\eta,\psi) = \exp{[x\eta - \psi(\eta) - h_{1}(x)]}
\end{equation}
where $h_{1}(x) = 0$,  $\eta = log{(\frac{q}{1-q})}$, and $\psi({\eta})=\log{(1+e^{\eta})}$, with $\eta$ denoting the natural parameter, $\psi(\eta)$ the log partition function and $x$ is the sufficient statistics associated with the distribution family. Using properties of the log partition function, we have the mean $\mu=E(x) = \nabla_\eta\psi = q$ and variance $\sigma^2=V(x) = \nabla^2_\eta\psi = q(1-q)$.

Similar to \cite{jiang2012small}, we now define a scaled version of the Bernoulli with natural parameter $\tilde{\eta} = \beta\eta$ and the log partition function $\tilde{\psi}(\tilde{\eta}) = \beta\psi(\frac{\tilde{\eta}}{\beta}) $, where $\beta >0$. Using the Lemma 3.1 of ~\cite{jiang2012small}, we can see that the mean $\tilde{\mu}$ and variance $\tilde{\sigma^2}$ of the scaled distribution $\tilde{p}(.)$ will be related to $\mu$ and $\sigma^2$ as   

\begin{align}
 \tilde{\mu} &= \nabla_{\tilde{\eta}} \tilde{\psi}(\tilde{\eta}) = \mu = q \\
 \tilde{\sigma}^2 &= \nabla_{\tilde{\eta}}^2 \tilde{\psi}(\tilde{\eta}) = \frac{\sigma^2}{\beta} = \frac{q(1-q)}{\beta}\\
\end{align}    

As discussed in ~\cite{banerjee2005clustering}, we can define a convex function $\phi$, that links Bernoulli to corresponding Bregman divergence. Let,
\begin{equation}
    \phi(x) = x\log{x} + (1-x)\log{(1-x)}
\end{equation}
Then, the Bregman divergence between a point $x$ and mean $\mu=q$ can be defined as:
\begin{align}
    d_\phi(x,\mu) &= \phi(x) - \phi(\mu) - (x - \mu)\nabla\phi(\mu) \\ 
    &=x\log{\frac{x}{q}} + (1-x)\log{\frac{1-x}{1-q}}
\end{align}
Using the Bregman divergence $d_\phi(x,\mu)$ defined above, the Bernoulli distribution can be expressed as

\begin{align*}
    P(x|\eta,\psi) &= \exp{[-d_\phi(x, \mu)]}f_\phi(x)
\end{align*}    

where $f_\phi(x) = \exp{(x\log{x} + (1-x)\log{(1-x)})}$

Now, we obtain the scaled version of the above likelihood by replacing $d_\phi(x,\mu)$ by $d_{\tilde{\phi}}(x,\tilde{\mu})$ , which in turn is $\beta \cdot d_{\phi}(x,\mu)$.
Denoting $\tilde{\phi} = \beta\phi$, the Bregman divergence representation of the scaled Bernoulli evaluates to be,
  \begin{align}
      \tilde{P}(x|\tilde{\eta},\tilde{\psi}) &= \tilde{P}(x|\tilde{\mu}) \\
      &= \exp{\{-d_{\tilde{\phi}}(x,\tilde{\mu})\}} \cdot f_{\tilde{\phi}}(x)\\
      &=\exp{\{-d_{\tilde{\phi}}(x,\mu)\}} \cdot f_{\tilde{\phi}}(x)
  \end{align}  
where, $f_{\tilde{\phi}}(x)=(f_\phi(x))^\beta$. With this representation of the scaled likelihood function established, we now discuss the MAP based asymptotics for the non-parametric model presented in the previous section.

\section{Applying SVA to LFRM}\label{Objective}

Having re-expressed the Bernoulli as a scaled Bernoulli, we are now in a position to derive SVA for the LFRM. For the LFRM, the joint posterior for the model will be 
\[
\mathcal{L}(\mathbf{Z},\mathbf{W}) = P(\mathbf{Z},\mathbf{W}|\mathbf{Y})  \propto P(\mathbf{Y}|\mathbf{Z},\mathbf{W}) P(\mathbf{Z}) P(\mathbf{W})
\]
We will be working with a loss function version of the objective, which can be written as the negative of the log posterior
\begin{align}
    -\log{\mathcal{L}(\mathbf{Z},\mathbf{W})} &= -\log{P(\mathbf{Y}|\mathbf{Z},\mathbf{W})}-\log{P(\mathbf{Z})} \\
    & -\log{P(\mathbf{W})} + \textrm{constant}
\end{align}    
Now, using the scaled Bernoulli representation, we get
\begin{align}
    P(\mathbf{Y}| \mathbf{Z}, \mathbf{W} )  &= \prod_{i, j=1}^{N} p_{ij}^{y_{ij}}(1-p_{ij})^{1-y_{ij}}\\
     & = \hspace{+2.0pt} \prod_{i, j =1}^{N} \hspace{-2.0pt} \exp[-\beta[y_{ij} \log{\frac{y_{ij}}{p_{ij}}} \\
     & + (1 - y_{ij})\log{(\frac{1-y_{ij}}{1-p_{ij}})}]] \\
    & \times \exp{[\beta[ y_{ij}\log{{y_{ij}}} + (1 - y_{ij})\log{(1-y_{ij})}]]}
\end{align}
This expression can be simplified to get
\begin{align*}
   -\log{P(\mathbf{Y}|\mathbf{Z},\mathbf{W})} = -\sum_{i=1}^{N}\sum_{j=1}^{N}\beta[y_{ij}\log{p_{ij}} \\
   + (1-y_{ij})\log{(1-p_{ij})}]
\end{align*}    

For the IBP prior term for $\mathbf{Z}$ (Eq.~\ref{eq:ibp}) we choose $\alpha = \exp{(-\beta \lambda^2)}$. The choice of this functional form is in line with the influence of $\alpha$ on the size of the binary latent representation size. Lower values of $\alpha$ promotes a smaller sized representation which is also the case with this form, in the limit of $\beta \to \infty$. This helps us avoid over-fitting of data to have the trivial latent feature representation of size $N$. $\lambda$ here is a hyperparameter, optimised by cross-validation. Substituting $\alpha$ for the expression of $p(\mathbf{Z})$ and simplifying we get
\begin{equation}
-\log{P(\mathbf{Z})} = K^{+}\beta\lambda^2 + \sum_{n=1}^{N} \frac{\exp{-(\beta \lambda^2)}}{n} + \textrm{constant(w.r.t. $\beta$)}
\end{equation}

Similarly, the negative log of prior for $\mathbf{W}$ is 
\begin{equation}
-\log{P(\mathbf{W})} = \sum_{k=1}^{K^{+}}\sum_{k'=1}^{K^{+}}\frac{w_{kk'}}{2\sigma^2} + \textrm{constant w.r.t $\sigma$}
\end{equation}

It is important to note here that the entire expression for $-\log{P(\mathbf{W})}$ is constant with respect to $\beta$. Therefore, the negative log posterior for $P(\mathbf{Z},\mathbf{W}|\mathbf{Y})$ can be written as
\begin{align*}
    -\log{\mathcal{L}(\mathbf{W},\mathbf{Z})}  &\propto -\log{P(\mathbf{Y}|\mathbf{W},\mathbf{Z})}-\log{P(\mathbf{Z})}-\log{P(\mathbf{W})}\\
    & = -\sum_{i=1}^{N}\sum_{j=1}^{N}\beta[y_{ij}\log{p_{ij}} \\
    &+ (1-y_{ij})\log{(1-p_{ij})}] + K^{+}\beta\lambda^2 \\
    &+ \sum_{n=1}^{N} \frac{\exp{-(\beta \lambda^2)}}{n} + \textrm{constant(w.r.t. $\beta$)}    
\end{align*}    

Dividing this equation by $\beta$ gives us
\begin{align*}
    -\frac{\log{\mathcal{L}(\mathbf{W}, \mathbf{Z})}}{\beta} &= K^{+}\lambda^2 + -\sum_{i=1}^{N}\sum_{j=1}^{N}[y_{ij}\log{p_{ij}} \\
    &+ (1-y_{ij})\log{(1-p_{ij})}] \\ &
    + \frac{\exp{-(\beta \lambda^2)}}{\beta}\sum_{n=1}^{N}\frac{1}{n} + O(\frac{1}{\beta})
\end{align*}    

Now, as $\beta \to \infty$, $O(\frac{1}{\beta})\to 0$ and $O(\frac{\exp{-(\beta \lambda^2)}}{\beta}) \to 0$. Thus we define the objective function, $\mathcal{Q}(\mathbf{W}, \mathbf{Z})$, which is to be minimized w.r.t. $\mathbf{W}$ and $\mathbf{Z}$, as
\small
\begin{align}
    \mathcal{Q}(\mathbf{W},\mathbf{Z}) \hspace{-2.0pt} &= \hspace{-2.0pt}\sum_{i=1}^{N}\sum_{j=1}^{N}[-y_{ij}\log{p_{ij}} \hspace{-2.0pt} -\hspace{2.5pt}(1\hspace{-2.0pt}-\hspace{-2.0pt}y_{ij})\log{(1-p_{ij})}] \hspace{-2.0pt} + \hspace{-2.0pt} C^{+}\\
    & = \sum_{i=1}^{N}\sum_{j=1}^{N}[-y_{ij}\log{\frac{p_{ij}}{(1-p_{ij})}} -\log{(1 - p_{ij})}] + C^{+}\\
    &= \hspace{-2.0pt} \sum_{i=1}^{N}\sum_{j=1}^{N}[-\hspace{-2.0pt}y_{ij}(\mathbf{z}_i^T\mathbf{W}\mathbf{z}_j) \hspace{-2.0pt} + \hspace{-2.0pt}\log{(1 \hspace{-2.0pt} + \hspace{-2.0pt} \exp{(\mathbf{z}_i^T\mathbf{W}\mathbf{z}_j)})}] \hspace{-2.0pt}+ \hspace{-2.0pt}C^{+}\\
\end{align}
\normalsize
\label{eq:finalobj}
 where, $p_{ij} = \sigma(\mathbf{z}_i^T\mathbf{W}\mathbf{z}_j)$ and $C^{+}=K^{+}\lambda^2$.
 
 Eq.~\ref{eq:finalobj} represents the MAP based equivalent objective for the nonparametric Bayesian LFRM~\cite{miller2009nonparametric}. Note that the objective consists of a sum of two component - the first component measures the fit to the data and the other component penalizing the number of latent features. The objective in Eq.~\ref{eq:finalobj} can be optimized w.r.t. $\mathbf{Z}$ and $\mathbf{W}$ using a variety of methods (both off-the-shelf as well as specialized optimizers). Also note that the objective is convex w.r.t. each $\mathbf{W}$ and $\mathbf{Z}$ (but not in both). In Sec.~\ref{sec:greedy}, we present a greedy algorithm to minimize this objective which alternates between optimizing $\mathbf{Z}$ and $\mathbf{W}$, and is guaranteed to reach a local minima of the objective.
 
 We would also like to note that the above formulation has striking similarity to the logistic regression loss function, where by using the trace trick, $\mathbf{z}_i^T\mathbf{W}\mathbf{z}_j=tr(\mathbf{W}^T\mathbf{z}_i \mathbf{z}_j^T)$. Here we can assume $\mathbf{z}_i \mathbf{z}_j^T$ to be the latent feature for each $y_{ij}$ term and $\mathbf{W}$ to be the model parameters. The trace term again can be expressed as a dot product of flattened matrices, making optimization of $\mathbf{W}$, for fixed $\mathbf{Z}$, exploit gradient based methods. Another important component of the objective is the penalty on the length of the latent representation $\mathbf{z}_i$. This has the benefit of not converging to the trivial case of $K^{+}=N$. An interesting aspect of the above objective is that it would stay valid for a wider variety of models with other link functions where the Bernoulli probabilities are not necessarily defined by a sigmoid $\sigma(\mathbf{z}_i^T\mathbf{W}\mathbf{z}_j)$ \cite{morup2011infinite}. 
\begin{table*}[!b]
\centering
 \begin{tabular}{ |c| c | c | c | c | c |}
 \hline
 \textbf{Method} & \textbf{Laz-Adv} & \textbf{Laz-Work} & \textbf{Laz-Fri} & \textbf{Protein230} & \textbf{NIPS234}  \\ 
   \hline 
 \textbf{MMSB} \cite{airoldi2008mixed}  & 0.813 & \textbf{0.844} & 0.846 & - & 0.871 \\  
 \hline
\textbf{HGP-EPM} \cite{zhou2015infinite} & - & - & - & 0.952 & 0.947 \\
 \hline
 \textbf{IRM} \cite{kemp2006learning}  & 0.796 & 0.826 & 0.821  & 0.934 & 0.948  \\ 
 \hline
  \textbf{LFRM-MCMC}\cite{miller2009nonparametric}  & 0.815 & 0.741 & 0.806 & 0.892 & 0.951 \\
 \hline
  \textbf{LFRM-SVA} (Ours) & \textbf{0.864} & 0.833 & \textbf{0.829} & \textbf{0.958} & \textbf{0.966} \\ 
  \hline
 \end{tabular}
 \caption{AUC-ROC evaluation for $50\%$-$50\%$ splits of Lazega-Lawyers networks and $80\%$-$20\%$ splits of Protein230 and NIPS234 datasets. In the table above, '-' denotes that the result for that baseline is not available for certain datasets}
  \label{table:other}
 \normalsize
\end{table*}
\section{Optimization}
\label{sec:greedy}
With the objective function in place, we now discuss the possible ways of achieving the optimal set of parameters $\mathbf{Z}$ and $\mathbf{W}$. The overall problem, under the small-variance asymptotic assumption gets reduced to solving the following optimization problem,
\begin{equation}
    \begin{aligned}
         &\minA_{\mathbf{W},\mathbf{Z}}\mathcal{Q}(\mathbf{W},\mathbf{Z})\\
         & \textrm{s.t. } \mathbf{z}_{i} \in \{0,1\}^{K^{+}} \quad \forall i=1\cdots N 
    \end{aligned}
\end{equation}
\subsection{Algorithm}

A simple starting point to optimize the above, would be to use a greedy strategy and optimize alternately with respect to $\mathbf{Z}$ and $\mathbf{W}$, similar in spirit to \cite{xu2015mad}. This would involve optimizing each $\mathbf{z}_i$ over all $2^{K^{+}}-1$ possible configurations, for fixed $\mathbf{W}$. We present a more greedy strategy, on the lines of the MAD-Bayes algorithm presented in \cite{broderick2013mad}, that first optimizes Eq.~\ref{eq:finalobj} for each element of $\mathbf{Z}$ and then with respect to $\mathbf{W}$. The complete algorithm K-LAFTER (Latent Feature learning on Relational data) is presented below, 
\begin{algorithm}[H]
\caption{K-LAFTER}\label{euclid}
\begin{algorithmic}[1]
\STATE  Initialize $K^{+}=1$ or larger, $\mathbf{Z}$ as a $N \times K^{+}$ matrix with $z_{ij}=1$ with probability 0.5 $\forall i=1\dots N, j=1\dots K^{+}$\\
\STATE Initialize $\mathbf{W}$ as $K^{+}\times K^{+}$ matrix with entries drawn from $\mathcal{N}(0,\sigma^2)$
\REPEAT
\STATE$\forall n, k$, Choose the optimal value(0 or 1) of each $z_{nk}$
\STATE optimize $\mathcal{Q}$ w.r.t. $\mathbf{W}$ for current $\mathbf{Z}$ \& $K^{+}$
\STATE Construct $\mathbf{Z}'$ from $\mathbf{Z}$ by adding a new feature by making $z_{n(K^{+}+1)}$ for a randomly initialized $n$ 
\STATE Augment $\mathbf{W}$ by drawing entries from $\mathcal{N}(0,\sigma^2)$ to form a $K^{+}+1$ dimensional square matrix $\mathbf{W}'$ 
\STATE optimize $\mathcal{Q}$ w.r.t. $\mathbf{W}'$ for current $\mathbf{Z}'$ \& $K^{+}+1$
\STATE optimize $\mathcal{Q}$ w.r.t. $\mathbf{Z}'$ for current $\mathbf{W}'$ \& $K^{+}$
\STATE If $(K^{+}+1, \mathbf{W}', \mathbf{Z}')$ lowers $\mathcal{Q}$ from $(K^{+},\mathbf{W},\mathbf{Z})$, replace latter with former
\UNTIL{convergence}
\end{algorithmic}
\end{algorithm}
The above algorithm can be sped-up further by caching values of the objective function by assuming each change of $z_{ij}$ from 0 to 1 (1 to 0) as an addition(subtraction) of a rank-1 elementary matrix, $\mathbf{M}$ with $m_{ij}=1$, 0 otherwise. 

The optimization w.r.t $\mathbf{W}$ can be performed by using $1^{st}$ order or $2^{nd}$ order batch/stochastic/co-ordinate gradient descent based methods, or using derivative-free methods that only use the objective function's value. In our implementation, we chose the latter.

\subsection{Proof of Local Convergence}
The proposed K-LAFTER algorithm converges to a local minima in finite number of iterations. We present a sketch of the proof for this. The first step of finding optimal $\mathbf{Z}$, for a fixed $\mathbf{W}$, always minimizes the objective because of its greedy nature. This is followed by the step of minimizing $\mathbf{W}$, for fixed $\mathbf{Z}$. As discussed in Sec.~\ref{Objective}, the objective is convex in $\mathbf{W}$ for a fixed $\mathbf{Z}$. Thus, this step realized by any order gradient descent style module, will lower the objective value. Next, while adding another dimension to latent representation, the choice is made greedily, choosing the one that has the lower objective value, thus moving closer to the local minima. 

\section{Related Work}
The small-variance asymptotics (SVA) has been leveraged recently to develop non-probabilistic counterparts for several nonparametric Bayesian latent variables models, and has resulted in fast deterministic inference algorithms for such models. Some of the notable examples include Dirichlet Process and hierarchical Dirichlet Process mixture models for clustering~\cite{kulis2011revisiting}, Indian Buffet Process based latent feature allocation for vector-valued data~\cite{broderick2013mad} with linear Gaussian observation model, the infinite Hidden Markov Model~\cite{roychowdhury2013small}, latent Dirichlet Allocation~\cite{jiang2017combinatorial}, etc. While these models are designed for i.i.d./sequential data, to the best of our knowledge, the SVA idea has not been applied to models for relational data, such as the latent feature relational model (LFRM), which is inherently a non-conjugate model, and for which the only known inference method is based on MCMC sampling~\cite{miller2009nonparametric}. 

Although not for LFRM, faster alternative to standard MCMC based inference have been developed for some other stochastic blockmodels, such as infinite relational model~\cite{kemp2006learning}, which assumes one-hot vector embedding for each node and the mixed-membership blockmodel~\cite{airoldi2008mixed}, which assumes a fractional membership of each node to multiple communities. These inference methods include methods based on online MCMC~\cite{li2016scalable} or online variational inference~\cite{gopalan2012scalable}. Applying these methods for LFRM is not straightforward. Online MCMC methods require carefully designed, model-specific derivations, which is further challenged by the discrete nature of the node embeddings. On the other hand, online variational inference to a model like LFRM is problematic due to the non-conjugacy of the LFRM~\cite{zhu2016max}. Our SVA based inference algorithm does not suffer from any of these issues. The final objective function has a simple form as a sum of a cross-entropy term and a regularizer that can be seen as penalizing large number of communities. The objective function can be optimized using a variety of inference methods, both batch and online. Moreover, although we assume the network data is be given in form of a binary matrix (presence/absence of an edge), other types of data can also be modeled (e.g., count-valued edges) by choosing an appropriate exponential family distribution for the likelihood.
\section{Experiments}

We now present experimental results of our SVA based inference algorithm for LFRM on various benchmark datasets. We compare our algorithm with MCMC based inference for LFRM, as well as with other state-of-the-art stochastic blockmodels on the link prediction accuracy. In addition, we also compare with MCMC in terms of link-prediction accuracy vs wall-clock time, to show that our algorithm attains much better link-prediction accuracies while taking a significantly shorter amount of time as compared to an MCMC sampler.

For our link-prediction experiments, we train all the models using $80\%$ of randomly chosen entries in the matrix $\mathbf{Y}$ data and the remaining $20\%$ of data is used to test the trained model. We  consider five random training-testing partitions for all datasets and report the average Area Under the Curve (AUC) of  the Receiver Operating Characteristic (ROC). Our model has only one free hyperparameter $\lambda$, which we tune using $k$-fold cross-validation technique on the training dataset. We would like to note that the performance of our algorithm is fairly insensitive to the exact choice of $\lambda$; in most cases, $\lambda = 0.5$ worked well. 

We initialize our $\text{K-LAFTER}$ algorithm (which we will refer to as LFRM-SVA in the rest of this section) with $K=1$. Initializing with larger $K$ is leads to slightly faster convergence. On all the datasets, our SVA based algorithm converged within 100 iterations if initialized with $K=1$, and in as few as 10 iterations if initialized with larger $K$ (e.g., $K=10$). The MCMC sampling based LFRM (referred to as LFRM-MCMC) was run for 1000 iterations with 500 burn-in and 500 collection iterations. We observed that the AUC scores of the MCMC based LFRM were fairly stable after these many iterations.

We report experimental results on the following benchmark datasets, also used in other prior work on LFRM~\cite{miller2009nonparametric} and other stochastic blockmodels~\cite{zhou2015infinite}.
 \begin{figure}[t]
     \centering
     \includegraphics[height=4cm,width=5.5cm]{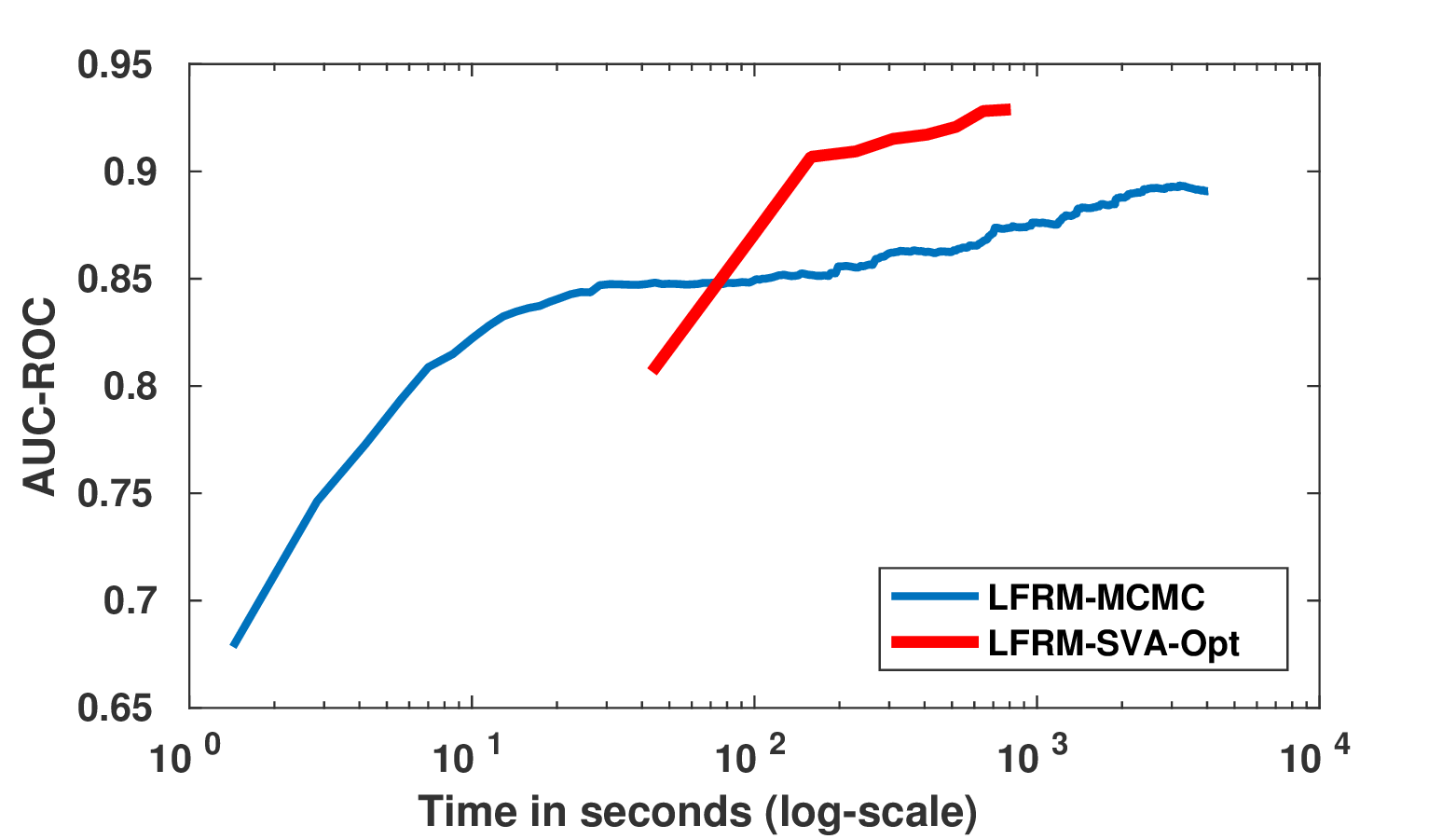}
     \vspace{-3pt}
     \caption{AUC vs wall-clock time comparison between LFRM with MCMC and LFRM with SVA on Protein230 data}
     \label{fig:protein_timing}
 \end{figure}
\begin{itemize}
\item \textbf{Lazega-Lawyers}\cite{lazega2001collegial}: This dataset constitutes of three small-scale networks and is based on corporate law partnership. The entities in these networks are lawyers and the relation predicates include symmetric relations like work based association, friendship association and the assymetric relation of advisory association.
\item \textbf{Protein230 Network}\cite{butland2005interaction}: This dataset consists of the interaction between 230 different proteins given in form of an adjacency matrix. The dataset has 595 edges. 

\item \textbf{NIPS234 Coauthor Network}\cite{miller2009nonparametric}:The NIPS234 network consists of 234 nodes with the relation describing the coauthorship of top 234 authors, by number of publications, in NIPS 1-17. 
\end{itemize}

We would like to note that we have chosen only moderate-sized datasets in our experiments so that it is feasible to run the MCMC sampler for LFRM for sufficiently large number of iterations, and do a fair comparison with our SVA based approach. The MCMC sampler does not scale easily to datasets with even a couple of thousand of nodes, while our SVA based algorithm does not face this issue.


Our experimental results on the link-prediction task for all the datasets are shown in Table~\ref{table:other}. As our experimental results show, LFRM-SVA attains much better link-prediction accuracies as compared to LFRM-MCMC, as well as various other state-of-the-art stochastic blockmodels, such as IRM, MMSB, HGP-EPM, etc. This can be attributed to the ability of our algorithm to search for a good solution (even though it is a point estimate) fairly quickly. In contrast, the MCMC based inference algorithm can take a long time to converge to a good solution.

 The convexity of the objective function in $\mathbf{W}$, for fixed $\mathbf{Z}$(step 5 and 8 in Algorithm \ref{euclid}), along with caching techniques for the greedy search of optimal $\mathbf{Z}$, while fixing $\mathbf{W}$ (step 4 and 9 in Algorithm \ref{euclid}), allows our proposed algorithm to scale to larger datasets and converge faster to higher AUC scores. This is also evident from Fig.~\ref{fig:protein_timing} where we compare the AUC vs wall-clock time for LFRM-MCMC and LFRM-SVA on Protein230 dataset. For this experiment, we initialized with $K=10$ and allowed both the algorithms to run until convergence of the AUC score. A similar experiment was also done for the NIPS234 dataset which yielded similar results, but skipped due to lack of space. The improvement in convergence speed can also be attributed to the fact that LFRM uses MCMC sampling based approach, where there are a fixed number of burn-in samples, followed by sampling from the approximated posterior. Here, usually the sampling subroutine becomes the bottleneck. The objective function formulated and the proposed algorithm are intended to put forward a scalable $k$-means style optimization trick and to drive small-variance asymptotics formulation of other Bayesian non-parametric models. While the datasets that have been discussed and evaluated on, have binary links present, we can easily extend the model to other datasets by an appropriate choice of the likelihood function and likewise formulating the objective.

  The latent feature representation of each entity learned by our model can also be used to perform a qualitative analysis, where each column of $\mathbf{Z}$ represents a latent community present in the network. An entity $i$, is a member of the community $k$, if $z_{ik}=1$ and not a part of it if $z_{ik}=0$. For the NIPS234 dataset, we choose communities with smaller number of members as they tend to represent a dense connection between the authors. We manually interpret the \emph{community name} based on the work of authors during the period from which the data was collected. Some of the communities are presented in \ref{table:authors}.  It is interesting to note that some authors like Thrun S, Bishop C etc. are inferred as belonging to multiple communities as the model allows overlapping communities.
\begin{table}[t]
\centering
\addtolength{\tabcolsep}{-1.0pt}
\begin{tabular}{ |p{2.4cm} |p{5.5cm} |}
 \hline
 \textbf{Community} & \textbf{Authors}  \\ 
  \hline 
 \textbf{Speech Processing} & SchmidBaur O, McNair A, Sloboda T, Woszczyna M, Doucet A, Hanson S \\  
  \hline 
 \textbf{Control and Robotics}  & Barto A, Sutton, Thrun S, Donoghue J, Burghard W \\ 
 \hline
\textbf{Computational Neuroscience}  & Stork D, Pawelzik K, Personnaz L, Dreyfus G, Pearlmutter B, Bishop C \\
 \hline
 \end{tabular}
  \caption{Communities of Authors obtained from their $\mathbf{Z}$ latent representations}
  \label{table:authors}
 \normalsize
\end{table}


\section{Conclusion}
We have presented a new inference algorithm for the latent feature relational model (LFRM) by applying the idea of small-variance asymptotics (SVA) to the LFRM. Our algorithm is simple to implement, faster than MCMC based inference for LFRM, and obtains comparable or better link-prediction accuracies on several benchmark datasets. Applying SVA to the LFRM results in an objective function that still retains the flavor of the nonparametric Bayesian flavor of LFRM (e.g., the ability to learn the number of communities), which opening doors to the possibility of choosing from a wide variety of optimization methods for learning the model parameters. Although we considered a greedy algorithm to optimize w.r.t. the binary latent feature matrix, recent advances in combinatorial optimization can also be leveraged to design other optimization algorithms for the objective. Other possible improvements include extending the optimization to work in an online setting or in a distributed setting, both of which are amenable under our SVA based setting. Finally, while our SVA based algorithm is a viable alternative for MCMC methods for doing inference for the LFRM, the fast point estimates produced by our method can also serve as good initializers for MCMC based inference for faster convergence since they rely critically on a good initializations.
\section*{Acknowledgements}
Gundeep Arora achnowledges the Research-I Foundation, IIT Kanpur. Piyush Rai acknowledges support from IBM Faculty Award, DST-SERB Early Career Research Award, Dr. Deep Singh and Daljeet Kaur Faculty Fellowship.

\appendix

\bibliographystyle{named}
\bibliography{ijcai18}

\end{document}